\documentclass{egpubl}
 
\JournalSubmission    
\usepackage[T1]{fontenc}
\usepackage{dfadobe}  

\usepackage{booktabs}
\usepackage{multirow}
\usepackage[table,xcdraw]{xcolor}
\usepackage{tcolorbox}
\usepackage[x11names, dvipsnames, table]{xcolor}    
\definecolor{blueboxcolor}{HTML}{57AAF2}
\definecolor{yellowboxcolor}{HTML}{F2B705}

\newtcolorbox{bluebox}[1][]{colframe=blueboxcolor, colback=blueboxcolor!10, boxrule=0.5mm, sharp corners, #1}
\newtcolorbox{yellowbox}[1][]{colframe=yellowboxcolor, colback=yellowboxcolor!10, boxrule=0.5mm, sharp corners, #1}
\usepackage{xspace}
\usepackage{amsmath,amssymb}

\newcommand{\para}[1]{\vspace{0.5ex}\noindent\textbf{#1}\hspace{1em}}

\usepackage{cite}  
\BibtexOrBiblatex

\electronicVersion
\PrintedOrElectronic

\usepackage{egweblnk} 

\title[High-fidelity Multi-view Normal Integration with Scale-encoded Neural Surface Representation]%
      {High-fidelity Multi-view Normal Integration with \\Scale-encoded Neural Surface Representation }

\author[Tongyu Yang, Heng Guo, Yasuyuki Matsushita, Fumio Okura, Yu Luo, Xin Fan]
{\parbox{\textwidth}{\centering Tongyu Yang\thanks{These authors contributed equally.}$^{1}$\orcid{0009-0001-0500-3687}, Heng Guo\footnotemark[1]\orcid{0000-0003-0047-3927}$^{2,4}$, Yasuyuki Matsushita$^3$\orcid{0000-0002-1935-4752}, Fumio Okura$^3$\orcid{0000-0001-7595-1300}, Yu Luo$^1$, Xin Fan$^1$\orcid{0000-0002-8991-4188} 
        }
        \\
{\parbox{\textwidth}{\centering     
    $^1$Dalian University of Technology, China\\
    $^2$Beijing University of Posts and Telecommunications, China\\
    $^3$The University of Osaka, Japan\\
    $^4$Beijing Key Laboratory of Multimodal Data Intelligent Perception and Governance
       }
}
}

\usepackage{xspace}

\usepackage{epsfig}
\usepackage{latexsym}
\usepackage{amsmath}
\usepackage{amssymb}
\usepackage{enumerate}
\usepackage{gensymb}
\usepackage{mathtools}
\usepackage{booktabs}
\usepackage{multirow}
\usepackage{siunitx}
\usepackage{tabularx}
\usepackage{arydshln}
\usepackage[nolist,nohyperlinks]{acronym}
\usepackage[acronym]{glossaries}
\usepackage[T1]{fontenc}
\usepackage{varioref}

\newacronym{PDEs}{PDEs}{partial differential equations}
\newacronym{made}{MADE}{mean absolute depth error}
\newacronym{GT}{GT}{ground-truth}

\usepackage{xspace}

\newcommand{\fref}[1]{Fig.~\ref{#1}}
\newcommand{\Fref}[1]{Figure~\ref{#1}}
\newcommand{\sref}[1]{Sec.~\ref{#1}}
\newcommand{\Sref}[1]{Section~\ref{#1}}
\newcommand{\eg}{\emph{e.g.}\xspace}
\newcommand{\ie}{\emph{i.e.}\xspace}

\usepackage{xcolor}
\newcounter{todos}
\AtEndDocument{\ifnum\value{todos}>0 \PackageWarning{TODOS}{There are \arabic{todos} todos left in this paper! Fix them before submitting the paper!} \fi}

\newcommand{\V}[1]{\ensuremath{\mathbf{#1}}}
\newcommand{\norm}[1]{\lVert#1\rVert}

\newcommand{\point}{\ensuremath{\V{p}}\xspace}
\newcommand{\cameraCenter}{\ensuremath{\V{o}}\xspace}
\newcommand{\rayDir}{\ensuremath{\V{v}}\xspace}

\newcommand{\R}{\ensuremath{\mathbb{R}}\xspace}
\newcommand{\radiusImage}{\ensuremath{\dot{r}}\xspace}
\newcommand{\radiusSpace}{\ensuremath{s}\xspace}
\newcommand{\focalLength}{\ensuremath{F}\xspace}

\newcommand{\argmax}{\mathop{\mathrm{argmax}}\limits}

\newcommand{\petneus}{PET-NeuS~\cite{wang2023petneus}\xspace}
\newcommand{\neus}{NeuS~\cite{wang2021neus}\xspace}
\newcommand{\sn}{SuperNormal~\cite{supernormal2024cao}\xspace}
\newcommand{\trimiprf}{Tri-MipRF~\cite{hu2023Tri-MipRF}\xspace}

\newcommand{\nerf}{NeRF~\cite{mildenhall2020nerf}\xspace}
\newcommand{\lodneus}{LoD-NeuS~\cite{zhuang2023anti}\xspace}
\newcommand{\rnbneus}{RNb-NeuS~\cite{Brument24}\xspace}

\newcommand{\analyticsplatting}{Analytic-Splatting~\cite{liang2024analyticsplatting}\xspace}
\newcommand{\mipsplatting}{Mip-Splatting~\cite{Yu2024MipSplatting}\xspace}
\newcommand{\ripnerf}{Rip-NeRF~\cite{liu2024ripnerf}\xspace}

\begin{document}


\maketitle
\begin{abstract}
Previous multi-view normal integration methods typically sample a single ray per pixel, without considering the spatial area covered by each pixel, which varies with camera intrinsics and the camera-to-object distance. Consequently, when the target object is captured at different distances, the normals at corresponding pixels may differ across views.
This multi-view surface normal inconsistency results in the blurring of high-frequency details in the reconstructed surface. To address this issue, we propose a scale-encoded neural surface representation that incorporates the pixel coverage area into the neural representation. By associating each 3D point with a spatial scale and calculating its normal from a hybrid grid-based encoding, our method effectively represents multi-scale surface normals captured at varying distances. Furthermore, to enable scale-aware surface reconstruction, we introduce a mesh extraction module that assigns an optimal local scale to each vertex based on the training observations. Experimental results demonstrate that our approach consistently yields high-fidelity surface reconstruction from normals observed at varying distances, outperforming existing multi-view normal integration methods.

\printccsdesc   
\end{abstract}  
\section{Introduction}
Detailed 3D surface reconstruction is essential for a wide range of applications, including 3D printing, virtual reality, and cultural heritage preservation~\cite{kim2025multiview, lv2024tansr}. Multi-view normal integration is a fundamental method for 3D reconstruction, enabling detailed surface geometry recovery given multi-view normal maps and camera poses, typically as a subsequent step to photometric stereo (PS)~\cite{Li_Zhou_Wu_Shi_Diao_Tan_2020, Wang_Ren_Guo_2023_ICCV}.

\begin{figure*}[htb]
	\centering
	\includegraphics[width=\textwidth]{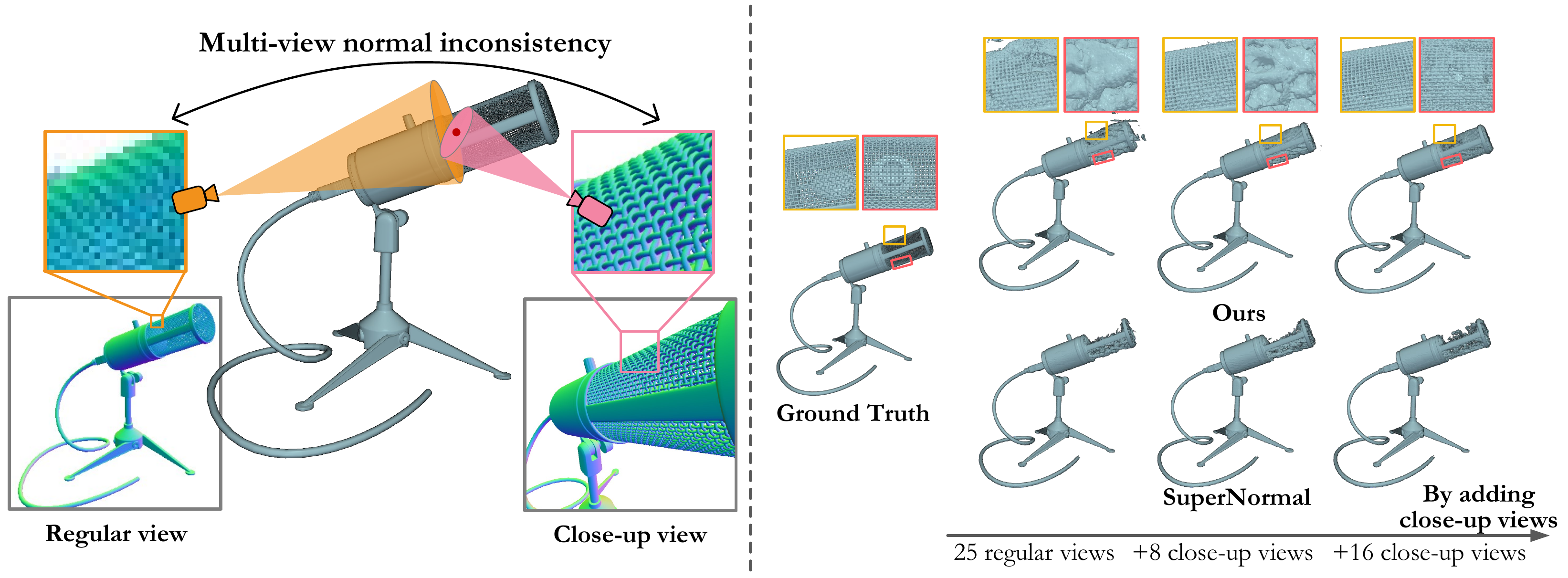}
	\caption{(a) Problem setup: Given surface normals captured at varying scales, the goal is to recover detailed 3D surfaces while addressing multi-view inconsistencies caused by scale variations. (b) Comparison with Existing Method~\cite{supernormal2024cao}: Our approach can handle surface normal inputs at multiple scales (close-up and regular), enabling the recovery of fine surface details, particularly in high-frequency regions. }
	\label{fig:teaser}
\end{figure*}

Existing multi-view normal integration methods~\cite{supernormal2024cao, Brument24} demonstrate a strong ability to represent high-frequency signals. They adopt a point-sampling strategy~\cite{mueller2022instant, neus2}, where each pixel is treated as a single point without accounting for the actual area it covers (which is determined by camera intrinsics and the camera-to-object distance). Consequently, they generally assume that the images are captured under the same camera-to-object distance with consistent intrinsics, so that the spatial resolution (i.e., the area size corresponding to each pixel) of surface normals across different views is roughly comparable. However, when the camera distances vary across views, the same region may correspond to different 2D  normal distributions, leading to scale-induced inconsistencies among multi-view observations. This motivates the need to explicitly incorporate scale information in the shape representation for high-fidelity 3D reconstruction.


In real-world image acquisition, high-frequency details are often concentrated in local regions. To capture them effectively, close-up shots are employed for detail-rich areas, while long shots are used to record the global geometry. \Fref{fig:teaser} (a) provides such an example: long shots capture the overall contour of the object as regular views, whereas close-up views reveal the fine structures of the microphone head. When zooming in on the normals of the same surface region across different views, noticeable inconsistencies emerge, as the recorded pixel normals correspond to different spatial resolutions of the same underlying surface area.

Existing multi-scale representation-based novel view synthesis (NVS) methods~\cite{barron2021mipnerf,hu2023Tri-MipRF} that take RGB images as input have successfully alleviated the multi-view inconsistencies caused by scale variations. For example, the integrated positional encoding (IPE) encoding in Mip-NeRF~\cite{barron2021mipnerf} approximates frustums with Gaussian distributions. However, these methods~\cite{barron2023zipnerf, liu2024ripnerf} are primarily designed to improve 2D NVS quality at arbitrary scales, rather than to ensure the geometric accuracy of 3D reconstruction, which involves a neural surface-constrained representation followed by mesh extraction. In particular, they fail to achieve a scale-aware mesh from a surface-constrained representation, since the scale of each spatial point is no longer determined from 2D observations. Adopting a single global scale for mesh extraction can lead to suboptimal outcomes: using a large scale tends to over-smooth fine local details captured by multi-scale observations, whereas using a small scale often produces unreliable meshes due to insufficient observation in most regions. This problem becomes particularly pronounced when the training dataset contains images captured from varying distances. To achieve high-quality geometric reconstruction, it is crucial to assign to each spatial location the most reliable local scale during mesh extraction so that fine details revealed by close-up views can be preserved. 

Motivated by the need for scale-aware reconstruction, we introduce a scale-encoded neural surface representation (SNSR) that addresses multi-view normal inconsistency caused by scale variations in the normal maps, enabling scale handling in both multi-view normal integration optimization and mesh extraction. It allows for high-fidelity multi-view normal integration across both close-up and global-shot surface normals. \Fref{fig:teaser} (b) illustrates the overall geometry of the microphone can be reconstructed with only regular views, but fine details are missing. As close-up views are gradually added, more detailed structures of the microphone emerge. In contrast, since the details among normals at different scales are blurred by point-based methods such as SuperNormal~\cite{supernormal2024cao}, they fail to effectively capture these fine-grained local structures.


Specifically, to handle multiscale surface normal input, SNSR encodes scale associated with each position into features. This approach models the surface normal as a function of scale, allowing each position to extract features at various scales. To achieve this, we sample spheres instead of points and incorporate each sphere's scale along the ray as an additional input. Furthermore, since incorporating scale results in a sparser representation, we introduce a cross-scale regularization (CSR) loss to regularize signed distance function (SDF) values across scales. Finally, to determine the scale of each point during surface extraction, we introduce an SNSR-based mesh extraction module (SMEM), which leverages the learned SNSR feature map to infer the well-trained scale corresponding to each position, enabling scale-aware mesh reconstruction. Our contributions are summarized as follows:
\begin{itemize}
    \item We propose a scale-encoded neural surface representation (SNSR) to resolve surface normal inconsistencies under large variations in spatial resolution;
    \item We present an SNSR-based mesh extraction module (SMEM) that enables scale-aware mesh extraction while preserving fine-grained geometric details;
    \item Experiments show that our method effectively handles multi-scale surface normals, enabling detailed reconstruction.
\end{itemize}

\label{sec:intro}

\section{Related Works}
This work focuses on multi-scale representation-based normal integration from multi-view normal maps. In the following, \sref{subsec: Surface normal integration} reviews related works on single-view and multi-view normal integration methods. \Sref{subsec:Multiscale 3D neural representation} surveys multi-scale representation methods for arbitrary-resolution rendering based on different underlying forms.

\subsection{Surface normal integration}
\label{subsec: Surface normal integration}
Normal integration refers to reconstructing a surface from a given normal field by solving for a scalar function whose gradient is consistent with the observed normals~\cite{heep2024screen-space-meshing,queau2018normal}.

In the single-view setting, the task reduces to integrating a single normal map. Classical approaches either formulate the problem as solving a Poisson equation~\cite{horn1986} or directly discretize the energy functional~\cite{durou2007integration}. Both cases lead to a linear system of equations. To tackle blurred depth discontinuities in normal integration, Cao \textit{et al}.~\cite{cao2022bilateral} assume the surface is one-sided differentiable everywhere in the horizontal and vertical directions, and propose a bilaterally weighted normal integration functional to preserve discontinuities.
More recent works focus on improving efficiency by moving beyond regular pixel grids, for instance, through adaptive meshing~\cite{heep2024screen-space-meshing} or mesh decimation strategies~\cite{heep2025feature}, while also preserving fine geometric details.

To extend normal integration to multiple views, recent approaches adopt implicit neural representations and differentiable rendering. SuperNormal~\cite{supernormal2024cao} employs directional finite differences and patch-based ray marching to approximate signed distance function (SDF) gradients, aiming for fine-grained surface details and efficiency. RNb-NeuS~\cite{Brument24} re-parameterizes reflectance and normals as radiance vectors under varying illumination, targeting improved multi-view reconstruction. PMNI~\cite{pmni} utilizes multi-view surface normals as geometric cues, jointly estimating camera poses and surface geometry. It improves reconstruction quality on reflective and textureless surfaces without requiring reliable initial poses.

However, these multi-view normal integration methods typically assume that normal maps are captured from roughly a constant distance from the object. When images are captured from varying distances, these methods treat each pixel as a single point instead of an area, which leads to ambiguity when different spatial resolutions are present at the same position, significantly limiting their performance.

\subsection{Multiscale 3D neural representation}    
\label{subsec:Multiscale 3D neural representation}
Multiscale representation enables models to render high-quality images at arbitrary scales. Achieving this requires explicitly accounting for the spatial resolution of pixels during rendering to ensure geometric and appearance consistency across different scales. Early point-sampled models (\eg, NeRF~\cite{mildenhall2020nerf}) do not explicitly model spatial resolution, which makes them prone to performance degradation when there is a mismatch between training and testing resolutions, thus limiting their ability to support stable cross-scale rendering. To address this issue, a variety of multiscale representation approaches have been proposed, which can be broadly categorized into three types: NeRF-based representations, SDF-based representations, and 3DGS-based representations.

\para{NeRF-based representation.} Mip-NeRF~\cite{barron2021mipnerf} extends \nerf by casting cones instead of rays, and uses 3D Gaussians to approximate the conical frustums of each pixel as the integrated positional encoding (IPE). To prevent aliasing in hybrid representation models~\cite{mueller2022instant}, Zip-NeRF~\cite{barron2023zipnerf} adopts a multi-sampling-based strategy to approximate the average NGP~\cite{mueller2022instant} feature over a conical frustum. By comparison, \trimiprf adopts a pre-filtering approach by emitting cones from the camera to the image plane, sampling spheres within each cone, and using TriMip encoding to represent each sphere. \ripnerf further improves \trimiprf with isotropic area sampling.

\para{SDF-based representation.} Despite providing implicit neural representations for NVS, NeRF-based methods struggle to extract high-quality surfaces due to soft density. \neus and its variants~\cite{wang2023petneus, wang2022hf} introduce neural SDFs for smoother and more detailed reconstructions, yet ray sampling still limits adaptive capture of varying detail across regions. To enable anti-aliased rendering and recover delicate geometric features, \lodneus aggregates spatial features via multi-convolved featurization within a conical frustum along a ray. RING-NeRF~\cite{ringnerf} introduces two key inductive biases: a continuous multi-scale representation and a decoder latent space invariant to both spatial location and scale, addressing NeRF’s common challenges such as robustness, handling unbounded scenes, and adaptive NVS.

\para{3DGS-based representation.} Currently, 3D Gaussian Splatting (3DGS)~\cite{kerbl3Dgaussians, gsoctree} is notable for its real-time rendering. However, a strong artifact is observed when changing the sampling rate. \analyticsplatting addresses aliasing by expanding each pixel integration area, enhancing sensitivity to spatial resolution variations of pixels across resolutions in 3DGS~\cite{kerbl3Dgaussians}. \mipsplatting uses a 3D smoothing filter combined with a 2D Mip filter to reduce high-frequency artifacts. Due to the discrete nature and the multi-view inconsistent behavior of 3D Gaussians, 3DGS~\cite{kerbl3Dgaussians} is difficult to directly apply for surface reconstruction. Some works~\cite{zhu2025gaussian, 2dgs} have explored surface reconstruction based on 3DGS representation, but they do not leverage a multi-scale representation.


The above-mentioned methods primarily focus on NVS, where the scale can be directly inferred from 2D observations. However, they do not consider scale during mesh extraction, where the optimal local scale cannot be directly determined. In contrast, aiming at high-quality 3D reconstruction, we propose a multi-scale representation and extend it with SMEM, enabling scale-aware mesh extraction and the recovery of high-fidelity geometric details.

\begin{figure*}[t] 
    \centering
    \includegraphics[width=\linewidth]{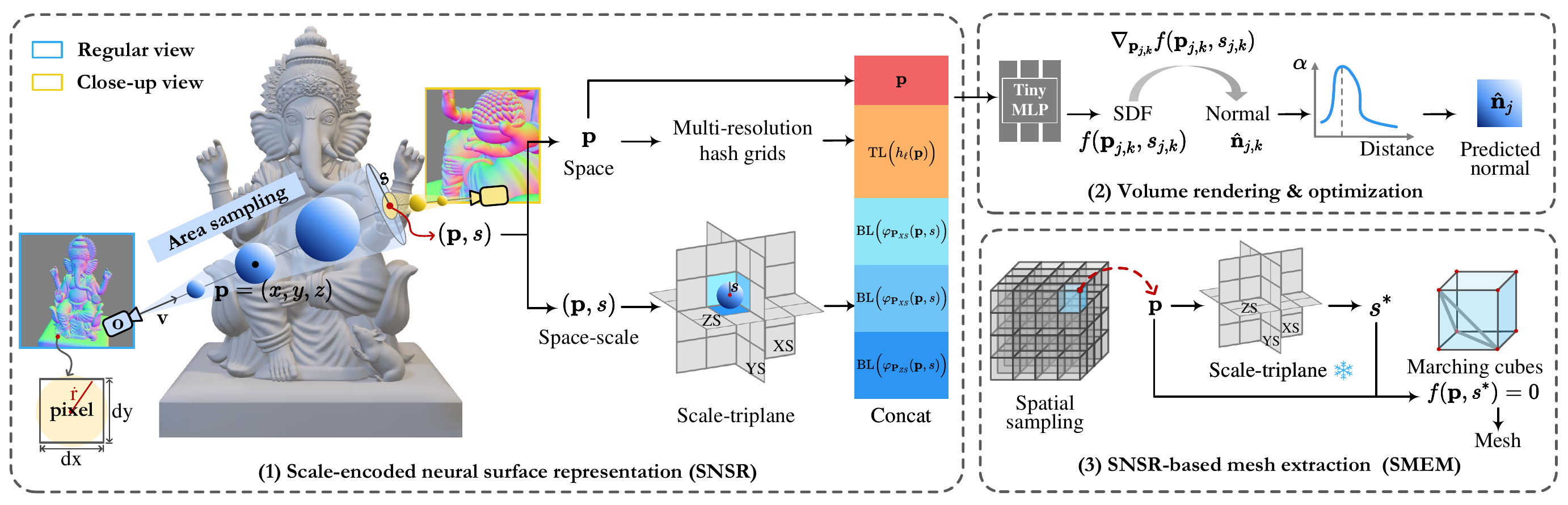}\vspace{-0mm}
    \caption{Method overview. (1) We sample inscribed spheres within the cone, each with position $\point$ and scale $s$, which are jointly encoded via hash grids and a scale-triplane. Interpolated features are concatenated and fed into a tiny MLP to output SDF and surface normal. (2) Pixel normals are computed via volume rendering.  (3) SMEM enables mesh extraction at the optimal scale.
    }
    \label{fig:method}
\end{figure*}

\section{Proposed Method}
\Fref{fig:method} shows an overview of our method. For multi-view normal maps captured at varying distances, we first adopt a sphere-based area sampling strategy, representing each sampled region by its center position and radius.  These position and scale parameters are then fed into hash grids and the scale-triplane module to obtain multi-scale feature representations. Next, surface normals along each ray are computed via volume rendering. The representation is optimized by a composite loss, including our proposed $\mathcal{L}_\text{CSR}$. During mesh extraction, we further introduce SMEM, an adaptive scale selection strategy to determine the optimal training-time scale for each vertex, enabling high-fidelity mesh reconstruction. 

\subsection{Scale-encoded neural surface representation}
\label{subsec:Sampling}

Given two types of normal maps captured at different distances, as shown in Fig.~\ref{fig:method}: regular view captures the overall shape of the object, and close-up view focuses on high-frequency details. The observation scales associated with these two views differ significantly, and neglecting this difference can lead to multi-view normal inconsistencies during optimization. 

To represent positions $\point=(x, y, z) \in \R^3$ as areas rather than individual points, we follow \trimiprf, casting a cone from the camera center \cameraCenter along the direction \rayDir and sampling inscribed spheres within the cone. The center position of each sampled sphere is given by:
\begin{equation}
    \point = \cameraCenter + t\rayDir, 
\end{equation}
where $t$ denotes the sampling distance along the ray. The radius \radiusSpace of each sphere encodes the scale of the corresponding point, which can be calculated as~\cite{hu2023Tri-MipRF}:
\begin{equation}
\radiusSpace = \frac{\norm{\point - \cameraCenter}_2 \cdot \focalLength \cdot \radiusImage}{\norm{\mathbf{v}}_2 \cdot \sqrt{\left(\sqrt{\norm{\mathbf{v}}_2^2 - \focalLength^2} - \radiusImage\right)^2 + \focalLength^2}},
\end{equation}
where $F$ denotes the focal length, and $\radiusImage$ is the scale of the disc on the image plane, computed as
\begin{equation}
\radiusImage = \sqrt{\frac{dy \cdot dx}{\pi}},
\end{equation}
where $dx$ and $dy$ are the width and height of the pixel in world coordinates that can be calculated from the camera intrinsics. 

\begin{figure}[t] 
    \centering
    \includegraphics[width=\linewidth]{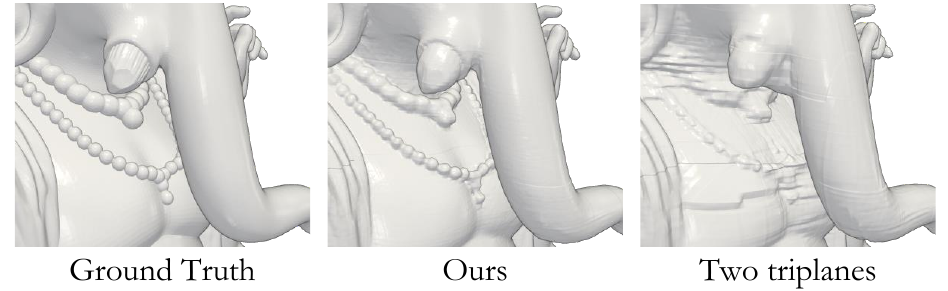}\vspace{-0mm}
    \caption{Comparison between our hybrid representation and two triplanes. Both methods use multi-resolution grids with the same finest spatial resolution.  }
    \label{fig:triplane}
\end{figure}

After obtaining the scale, each point can be represented as a 4D coordinate $(\point,s)$. Two straightforward approaches are possible to encode this coordinate: a full 4D hash grid which replicates a 3D hash grid at each scale, or a decomposition into two triplanes—three planes $\mathbf{P}_{XY}, \mathbf{P}_{XZ}, \mathbf{P}_{YZ}$ for spatial dimensions and three planes $\mathbf{P}_{XS}, \mathbf{P}_{YS}, \mathbf{P}_{ZS}$ for scale. However, the former leads to an excessively large model when scales vary~\cite{kplanes_2023}, while the latter struggles to capture high-frequency spatial details (\fref{fig:triplane}). 

To leverage the strengths of both representations, we adopt a hybrid multi-scale neural representation: a multi-resolution hash grid encodes spatial positions, while a scale-triplane $\mathbf{P}{XS}, \mathbf{P}{YS}, \mathbf{P}_{ZS}$ encodes the joint position–scale space. This design provides continuous representation across arbitrary scales, explicitly links scale to surface normals, and reduces memory or improves detail representation compared to a full 4D grid or dual-triplane approach. \Fref{fig:triplane} illustrates that our method recovers finer details and requires less time (40 minutes) for the same number of iterations, compared to 120 minutes for the two triplanes representation.

In particular, Fig.~\ref{fig:method} illustrates that we input $\point$ into a multi-resolution hash encoding to obtain spatial features, while $(\point,s)$ are fed into a single-resolution scale-triplane to capture scale-dependent features. Feature vectors are obtained by linearly interpolating according to the position and scale from the hash encoding and triplane, respectively. Then, we concatenate these features as the input to a tiny MLP, which decodes them into SDF values $f$: 

\begin{equation} \label{eq:sdf_proj_c}
f(\mathbf{p}, s) = \text{MLP} \left\{
\begin{aligned}
    &\point, \\
    &\text{TL}\Big(h_\ell(\point)\Big), \ell = \{1 \dots L\}, \\
    &\text{BL}\Big(
        \varphi_{c}(\point, s)\Big), c = \{\textbf{P}_{XS}, \textbf{P}_{YS}, \textbf{P}_{ZS}\}
\end{aligned}
\right\}_{\text{concat}},
\end{equation}
where $\text{TL}$ and $\text{BL}$ denote the trilinear and bilinear interpolation operations, respectively, $h_\ell$ is the $\ell$-th level of hash grids, and $\varphi_c$ projects $(\mathbf{p}, s)$ onto the $c$-th plane.


\subsection{Volume rendering of surface normal}
After obtaining SDF values at sampled points, we compute surface normals along each ray using volume rendering, which allows supervision from the ground-truth normal maps to optimize our multi-scale representation. We adopt PyTorch’s automatic differentiation approach. For the $k$-th sample point along the $j$-th ray $\mathbf{p}_{k,j}$, its predicted normal $\hat{\mathbf{n}}_{k}$ is given by:
\begin{equation}
\hat{\mathbf{n}}_{j,k} =
\nabla_{\mathbf{p}_{j,k}} f(\mathbf{p}_{j,k}, s_{j,k}).
\end{equation}

Then, following \neus, we adopt an unbiased and occlusion-aware method to convert SDF samples into volume rendering opacities, allowing us to transform the signed distance values into densities:
\begin{equation}
    \alpha_{j,k} = \max\left(\frac{{\Phi_{a}}(f(\mathbf{p}_{j,k}, s_{j,k})-{\Phi_{a}}(f(\mathbf{p}_{j,k+1}, s_{j,k+1})} {\Phi_{a}(f(\mathbf{p}_{j,k}, s_{j,k})}  , 0 \right),
\end{equation}
where $\alpha_{j,k}$ is discrete opacity values, and $\Phi_a(\,\cdot\,) = \big(1 + e^{-a\,(\cdot\,)}\big)^{-1}$ is the sigmoid function with a trainable sharpness parameter $a$. 

Finally, the normal along the $j$-th ray $\hat{\mathbf{n}}_j$ is computed via volume rendering:
\begin{equation}
    \hat{\mathbf{n}}_j = \sum_{k=1}^{M} T_{j,k} \, \alpha_{j,k} \,  \hat{\mathbf{n}}_{j,k},
\quad \text{with} \quad
T_{j,k} = \prod_{l=1}^{k-1} (1 - \alpha_{j,l}),
\end{equation}
where $M$ is the number of the sampled points of each ray.

\subsection{Loss functions}
To optimize our multi-scale representation, we employ the following loss functions.

\para{Normal loss.} 
The normal loss encourages the estimated normals $\mathbf{\hat{n}}$ to align with the input normal vectors $\mathbf{n}$, thereby facilitating the recovery of accurate geometric details:
\begin{equation}\label{eq:normal loss}
 \mathcal{L}_{\text{normal}} = \frac{1}{N}\sum_{j=1}^{N} \left\| \mathbf{\hat{n}}_j - \mathbf{n}_j \right\|_1,
\end{equation}
where $N$ is the number of rays.

\para{Mask loss.} 
The mask term aligns the shape's silhouette with the input masks, serving as a boundary condition.
\begin{equation}
\mathcal{L}_{\text{mask}} 
= - \frac{1}{N} \sum_{j=1}^{N} \Big[\, m_j \log(\hat{m}_j) + (1 - m_j) \log(1 - \hat{m}_j) \,\Big],
\end{equation}
where $\hat{m}_j= \sum_{k=1}^{M} T_{j,k} \alpha_{j,k}$ denotes the accumulated opacity along the $j$-th ray,  and $m_j \in \{0,1\}$ denotes the ground-truth binary mask value at the $j$-th pixel.

\para{Eikonal loss.} 
Eikonal loss~\cite{Osher_Fedkiw_2003} constrains the gradient norm of the SDF to be close to 1 throughout the space, ensuring that the neural SDF remains a valid representation of the surface geometry.
\begin{equation}
 \mathcal{L_{\text{eikonal}}} = \frac{1}{NM}\sum_{j,k} \left( \| \nabla_{\mathbf{p}_{j,k}} f(\mathbf{p}_{j,k},s_{j,k}) \|_2 - 1 \right)^2.
\end{equation}

\para{Cross-Scale regularization loss.} 
The additional scale input makes our method more prone to overfitting the normal maps, resulting in high-quality normal maps but inaccurate depth estimations (as shown in Fig.~\ref{fig:sdf}, row 1, column 2 and 3). This, in turn, negatively impacts mesh generation. 
The issue arises because not all regions appear in close-up views, \ie, not all points have been observed at smaller scales. For these unobserved scales, there is no corresponding supervision for optimization. Therefore, we introduce an unsupervised Cross-Scale regularization term $\mathcal{L_{\text{CSR}}}$ to associate the SDF values at these scales with the reliably trained SDF, and prevent the model from over-fitting. Specifically, in each iteration, we randomly select $K$ sampled points and uniformly sample $S$ scales within the range between the minimum and maximum radius. It is worth noting that only a small number of points and scales are required to perform this regularization effectively.
\begin{equation}
\begin{aligned}
    \mathcal{L}_{\text{CSR}} &= \frac{1}{KS} \sum_{k=1}^{K} \sum_{i=1}^{S}
    \left\| f(\mathbf{p}_{k},s_{k,i}) - \mu_{k} \right\|_{2}^{2}, \\
    &\text{with} \quad
    \mu_{k}= \frac{1}{S} \sum_{i=1}^S f(\mathbf{p}_{k},s_{k,i}).
\end{aligned}
\end{equation}


\para{Overall loss.}
The overall loss function is given by:
\begin{equation}\label{eq:loss}
 \mathcal{L} = \mathcal{L}_{\text{normal}} + \mathcal{L}_{\text{mask}} + \mathcal{L}_{\text{eikonal}} + \lambda\mathcal{L}_{\text{CSR}}, 
\end{equation}
where $\lambda$ denotes the weight of $\mathcal{L}_{\text{CSR}}$. 
 
 \begin{figure*}[t]
    \centering
    \includegraphics[width=0.9\linewidth]{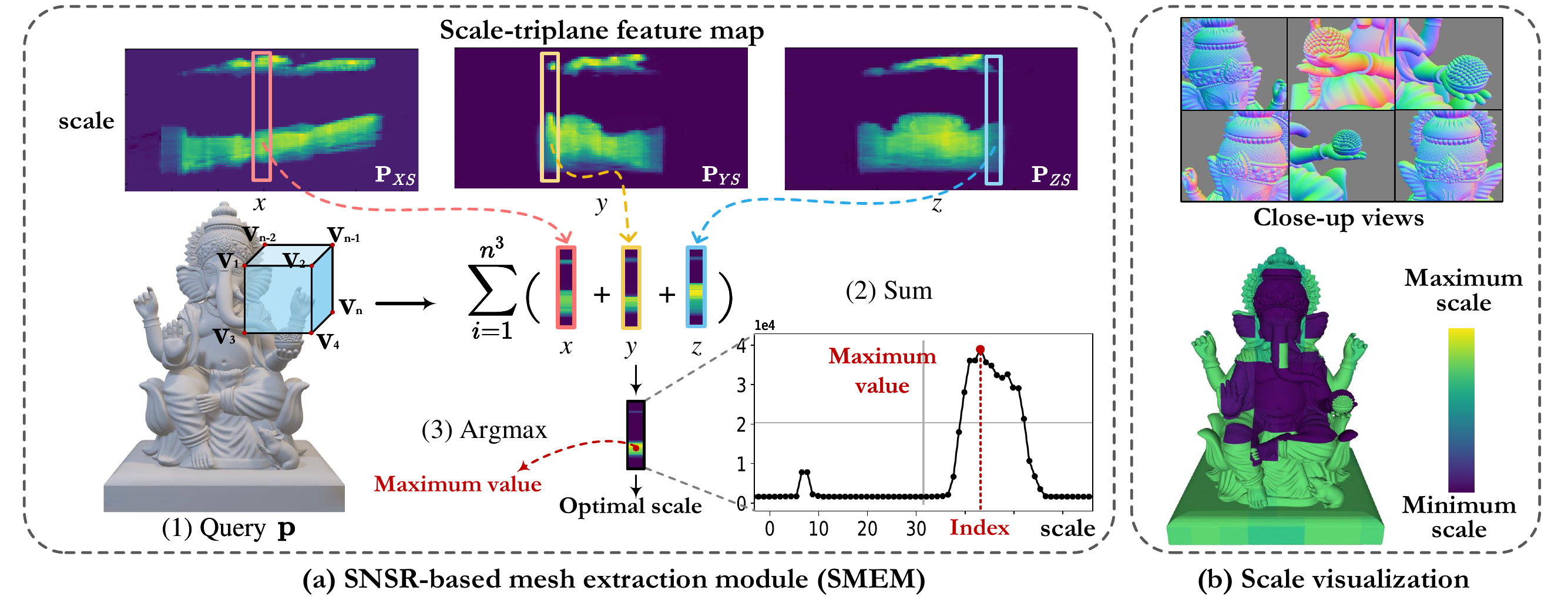}
    \caption{
    The flowchart of SMEM. (a) For each vertex within a cell, we query the scale-triplane feature map at its position $\point$, sum the features of $x$, $y$, and $z$ axes for all vertices within the cell, and select the scale with the maximum aggregated response as the cell’s final scale. (b) We visualize the scales on the extracted mesh, where the purple areas correspond to smaller scales, which is consistent with the observations from the close-up views. 
    }
    \label{fig:mesh}
\end{figure*}

\subsection{SNSR-based mesh extraction}
To extract mesh from the SDF, we define the surface $\mathcal{S} $ as the zero-level set of the SDF $f$:
\begin{equation}
 \mathcal{S} = \left\{ (\point,s) \in \mathbb{R}^4 \mid f(\point,s) = 0 \right\}.
\end{equation}

During training, each point is assigned an optimal scale based on its view, allowing the model to learn with the most suitable scale for each position. Ideally, these learned scales could guide mesh extraction to recover the most detailed geometry. However, since mesh extraction is independent of the observed images, the scale corresponding to the finest SDF at each spatial location cannot be directly determined, making it challenging to reconstruct a high-fidelity 3D mesh. To address this, we design an SNSR-based mesh extraction module (SMEM) to adaptively find the optimal scale for each vertex with the marching cubes algorithm~\cite{lorensen1998marching}.  

Since the optimal scale used for mesh extraction is derived from the training observations, we can naturally obtain the scale of each vertex from the trained scale-triplane feature map. We visualize this feature map in Fig.~\ref{fig:mesh} and observe that the maximum value on the triplane aligns with the well-trained scale. This phenomenon can be explained by the initialization of the triplane features with zeros: scales that are rarely observed tend to remain close to zero, whereas positions that are sufficiently constrained by multi-view observations accumulate stronger activations during training, resulting in higher feature values. Consequently, the maximum response over scales at a given position naturally indicates the scale at which the representation has been most effectively learned.

\Fref{fig:mesh} depicts the flowchart of our SMEM. Based on this observation, we first query each vertex on the three planes $\textbf{P}_{XS}, \textbf{P}_{YS}, \textbf{P}_{ZS}$ to obtain three sets of feature values corresponding to the $X$, $Y$, and $Z$ coordinates, covering all the scales involved in training for that vertex. Since the maximum feature responses along the three axes do not necessarily correspond to the same scale, while the training process consistently optimizes all three axes at a shared scale, we sum the three feature values to identify the scale that jointly maximizes their responses, which is then selected as the final scale.

Using this insight, we first query each vertex on the feature map to obtain three sets of values, each corresponding to $x$, $y$, and $z$ coordinates, respectively. Since each $(x,y,z)$ position corresponds to the same scale, we sum $xs$, $ys$, and $zs$ to yield one column of scales per vertex, from which we determine a single optimal scale. The final optimal scale for each vertex is selected by summing these columns and choosing the scale associated with the maximum value from the feature map. We visualize each vertex’s scale on the mesh, as shown in Fig.~\ref{fig:mesh} (b). Purple sections represent regions where a smaller scale is used for mesh extraction, corresponding to close-up view coverage, while other regions apply the scale associated with regular views.



Directly selecting the scale corresponding to the maximum feature value for each vertex may lead to surface discontinuities, as SDF vary significantly across different scales. To mitigate this issue, we partition the entire volume into units $n\times n\times n$ (with $n=64$), and assign a consistent scale to all vertices within each unit. The final scale $s^*$ for each unit volume is calculated by: 
\begin{equation}
s^* = \argmax_s\sum_{i=1}^{n^3} \sum_{c} \text{BL}\Big(
        \varphi_{c}(\point)\Big),c = \{\textbf{P}_{XS}, \textbf{P}_{YS}, \textbf{P}_{ZS}\}. \!
\end{equation}

\begin{figure*}[t]
	\centering
	\includegraphics[width=\linewidth]{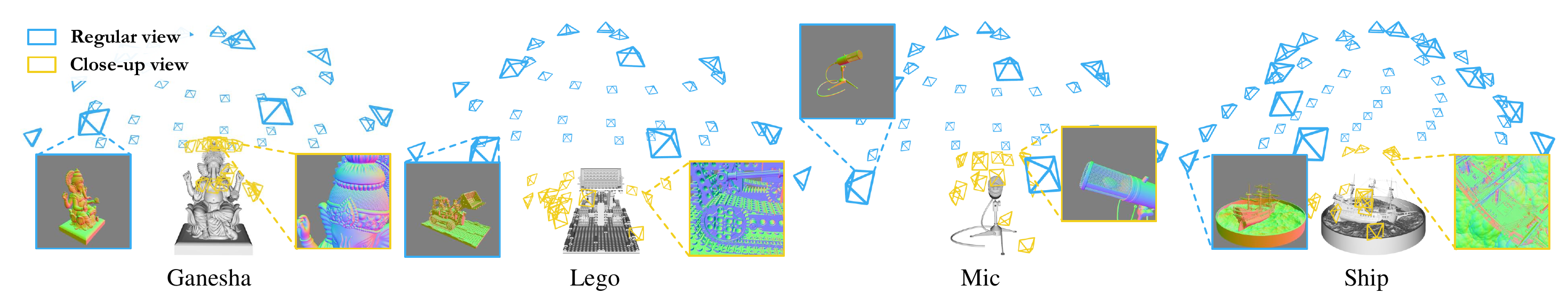}
	\caption{Setup of synthetic data generation. Blue and yellow camera poses denote the regular and close-up views, respectively. We provide example views for each object.} 
	\label{fig:synthetic_data}
\end{figure*}

\para{Implementation details.}
The input 3D coordinate $\mathbf{p}$ and scale $s$ are transformed through our scale-triplane encoding and hash encoding~\cite{mueller2022instant} into a $55$-dimensional feature vector ($3$ dimensions from position $\mathbf{p}$, $28$ dimensions from multi-resolution hash grids, and $24$ dimensions from Scale-triplane). The hyperparameter setting of hash grids is the same as \sn. Subsequently, a tiny multi-layer perceptron (MLP) with 64 units is used to decode the features, with ReLU activation applied to the single hidden layer. Additionally, the resolution of each plane in the triplane representation is $128 \times 32$, where $128$ and $32$ correspond to the resolution of the 3D coordinate  $\mathbf{p}$ and scale $s$, respectively. We randomly sample $4096$ rays per batch, and the total number of iterations is $70,000$. We use the Adam optimizer with an initial learning rate $5 \times 10^{-3}$. For each iteration, we use $128$ rays with $4096$ sampled points, and for $\mathcal{L}_\text{CSR}$, we randomly select $4096$ points from it, uniformly sample $128$ scales, and set $\lambda=4$

\section{Experiments}
In this section, we present a comparison between our method and existing normal integration approaches, and conduct an ablation study to demonstrate the effectiveness of our approach.

\subsection{Experimental settings}
\label{subsec:setting}

\para{Datasets.} 
To compare our method with existing multi-view normal integration approaches, we render synthetic datasets at a resolution of $800 \times 800$, containing four high-frequency objects selected from NeRF~\cite{mildenhall2020nerf} and Sketchfab~\cite{sketchfab}. For data generation, \textit{Setting I} consists of about $32$ normal maps rendered in Blender from a constant distance. \textit{Setting II} augments \textit{Setting I} with an additional with $8$ to $16$ close-up views that capture fine details which are challenging to observe in regular views. \Fref{fig:synthetic_data} illustrates the camera pose distributions used for normal rendering and examples of regular and close-up views for each object.
\textit{Setting I} highlights the benefit of close-up views, while \textit{Setting II} introduces larger multi-view normal inconsistencies. We also render RGB for training rgb-based methods. 

\begin{figure}[t]
	\centering
	\includegraphics[width=\linewidth]{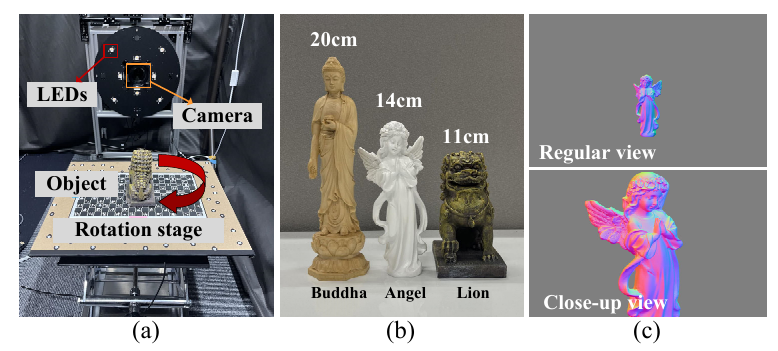}
	\caption{Setup of our real-world experiment. (a) The target object is placed on a checkerboard and imaged on a rotation stage under $12$ lighting conditions. (b) The objects range in height from $11$ cm to $20$ cm. (c) Normal maps estimated from SDM-UniPS~\cite{uni_ps}. }
	\label{fig:exp}
\end{figure}

%
For the real-world dataset, since ground-truth normal maps are unavailable, we follow the existing multi-view normal integration approach~\cite{supernormal2024cao}, which  applies SDM-UniPS~\cite{uni_ps} to estimate per-view surface normals from images captured under multiple illumination conditions. To this end, we capture images using the setup illustrated in \fref{fig:exp} (a). We select three objects with complex geometric shapes (\fref{fig:exp} (b)), each imaged from $26$ viewpoints ($24$ regular views and $2$ close-up views) under $12$ lighting conditions. Images are acquired using a Canon EOS RP at a resolution of $3360 \times 2240$. Camera parameters are calibrated with Metashape~\cite{agisoft}, foreground masks are generated with SAM~\cite{sam}.

\begin{table*}[tp]
\renewcommand{\arraystretch}{1.1} 
\caption{Quantitative evaluation on synthetic data. Red and orange cells indicate \colorbox[HTML]{FFB3B3}{the best} and \colorbox[HTML]{FFD9B3}{the second best} results, respectively. Results are measured in terms of MAE and CD. } 
\centering
\small
\begin{tabular}{cccccccccc}
\midrule
\multicolumn{1}{c}{{\color[HTML]{000000} }}                         & \multicolumn{4}{c}{Mean Angular Error $\downarrow$ (novel views)}                                                                                                   & \multicolumn{4}{c}{L2 Chamfer Distance $\downarrow$}                                                                                                                   &                                 \\ \cline{2-9}
\multicolumn{1}{c}{\multirow{-2}{*}{{\color[HTML]{000000} Method}}} & Ganesha                         & Lego                            & Mic                            & \multicolumn{1}{c}{Ship}                           & Ganesha                          & Lego                           & Mic                             & \multicolumn{1}{c}{Ship}                            & \multirow{-2}{*}{Runtime$\downarrow$}      \\ \midrule
\multicolumn{10}{l}{\textbf{Setting I, only   regular views:}}                                                                                                                                                                                                                                                                                                                                                               \\ \midrule
\multicolumn{1}{c}{\sn}                                     & 7.5770                          & 16.815                          & \cellcolor[HTML]{FFD9B3}13.551 & \multicolumn{1}{c}{\cellcolor[HTML]{FFD9B3}7.103}                          & 0.2577                          & 0.3476                         & \cellcolor[HTML]{FFD9B3}0.1899  & \multicolumn{1}{c}{\cellcolor[HTML]{FFD9B3}0.1516}  & \cellcolor[HTML]{FFD9B3}14 mins \\
\multicolumn{1}{c}{\rnbneus}                                        & 7.5810                          & 17.061                          & 14.655                         & \multicolumn{1}{c}{7.648}                          & \cellcolor[HTML]{FFD9B3}0.1809  & 0.5677                         & 0.6132                          & \multicolumn{1}{c}{0.2563}                          & 11 hrs                          \\
\multicolumn{1}{c}{\petneus}                                   & \cellcolor[HTML]{FFD9B3}6.7042                         & \cellcolor[HTML]{FFD9B3}13.549  & 18.818                         & \multicolumn{1}{c}{7.149}                          & 0.1837                          & \cellcolor[HTML]{FFD9B3}0.2028                         & 0.5415                          & \multicolumn{1}{c}{0.1634}                          & 26 hrs                          \\
\multicolumn{1}{c}{\trimiprf}                                   & - & -                          & -                         & \multicolumn{1}{c}-  & 0.5111                         & 0.7477 & 0.4906                          & \multicolumn{1}{c}{0.6730 }                          & \cellcolor[HTML]{FFB3B3}13 mins                          \\
\multicolumn{1}{c}{Ours}                                            & \cellcolor[HTML]{FFB3B3}4.1205 & \cellcolor[HTML]{FFB3B3}11.1419 & \cellcolor[HTML]{FFB3B3}8.5297 & \multicolumn{1}{c}{\cellcolor[HTML]{FFB3B3}4.7417} & \cellcolor[HTML]{FFB3B3}0.0617 & \cellcolor[HTML]{FFB3B3}0.1474 & \cellcolor[HTML]{FFB3B3}0.0812 & \multicolumn{1}{c}{\cellcolor[HTML]{FFB3B3}0.0704}  & 34 mins   \\ \midrule
\multicolumn{10}{l}{\textbf{Setting II, regular   + close-up views:}}                                                                                                                                                                                                                                                                                                                                                        \\ \midrule
\multicolumn{1}{c}{\sn}                                     & 6.7823                         & 12.263                          & \cellcolor[HTML]{FFD9B3}12.403 & \multicolumn{1}{c}{\cellcolor[HTML]{FFD9B3}6.664}  & 0.1606                            & 0.1768                         & \cellcolor[HTML]{FFD9B3}0.1601  & \multicolumn{1}{c}{\cellcolor[HTML]{FFD9B3}0.1405}  & \cellcolor[HTML]{FFD9B3}14 mins \\
\multicolumn{1}{c}{\rnbneus}                                        & 6.7518                         & 18.47                           & 14.679                         & \multicolumn{1}{c}{7.956}                          & \cellcolor[HTML]{FFD9B3}0.1474  & 0.3504                         & 0.8635                          & \multicolumn{1}{c}{0.3273}                          & 11 hrs                          \\
\multicolumn{1}{c}{\petneus}                                   & \cellcolor[HTML]{FFD9B3}6.0367                         & \cellcolor[HTML]{FFD9B3}11.7567                         & 14.778                         & \multicolumn{1}{c}{6.871}                          & 0.177                           & \cellcolor[HTML]{FFD9B3}0.1477 & 0.3927                          & \multicolumn{1}{c}{0.1437}                          & 26 hrs                          \\
\multicolumn{1}{c}{\trimiprf}                                   & - & - & -                         & \multicolumn{1}{c}{-}                          & 0.5608                          & 0.6576                         & 0.7784                           & \multicolumn{1}{c}{0.6947}                          & \cellcolor[HTML]{FFB3B3}13 mins                          \\
\multicolumn{1}{c}{Ours}                                            & \cellcolor[HTML]{FFB3B3}3.7028 & \cellcolor[HTML]{FFB3B3}8.8386  & \cellcolor[HTML]{FFB3B3}7.7164 & \multicolumn{1}{c}{\cellcolor[HTML]{FFB3B3}4.5801} & \cellcolor[HTML]{FFB3B3}0.0483 & \cellcolor[HTML]{FFB3B3}0.1131 & \cellcolor[HTML]{FFB3B3}0.0809  & \multicolumn{1}{c}{\cellcolor[HTML]{FFB3B3}0.0684} & 34 mins   \\ \midrule
\end{tabular}
\label{tab:compare}
\end{table*}

\begin{figure*}[tp]
    \centering
    \includegraphics[width=0.83\linewidth]{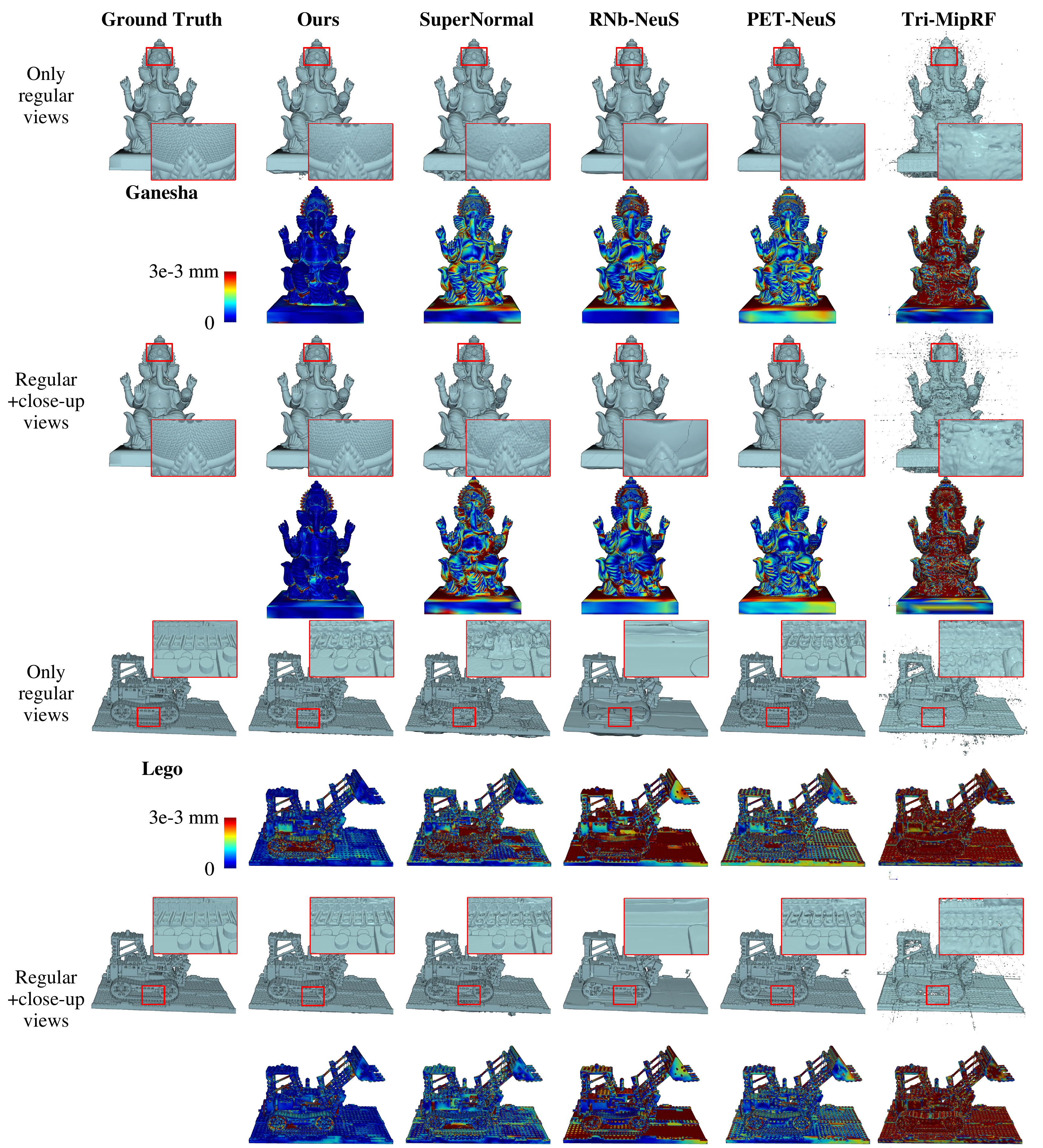}
    \caption{Comparison between our method and baselines on synthetic data. 
    } 
    \label{fig:compare}
    \vspace{-1em} 
\end{figure*}

\begin{table}[t]
\renewcommand{\arraystretch}{1.1} 
\caption{Ablation study on SNSR and $\mathcal{L}_\text{CSR}$ on synthetic data. \textbf{Bold text} indicates the best result.   } 
\centering
\small
\begin{tabular}{
c c c c c }
\midrule                   & \multicolumn{2}{c}{Setting I}  & \multicolumn{2}{c}{Setting II}                                                           \\ \cline{2-5} 
\multirow{-2}{*} & MAE $\downarrow$                                                       & CD $\downarrow$                                                          & MAE $\downarrow$                                                      & CD $\downarrow$                                                         \\ \midrule
w/o SNSR                                                     & 4.622                                                      & 0.1119                                                        & 5.342                                                     & 0.097                                                      \\
w/o $\mathcal{L}_\text{CSR}$                                                     & 4.455                                                      & 0.0687                                                       & 4.443                                                     & 0.068                                                        \\
Ours                                                              & \textbf{4.121} & \multicolumn{1}{l}{\textbf{0.0617}} & \textbf{3.703} & \multicolumn{1}{l}{\textbf{0.048}} \\ \midrule
\end{tabular}
\label{tab:ablation}
\end{table}

\begin{table*}[tbp]
\small
\renewcommand{\arraystretch}{1.1} 
\setlength{\tabcolsep}{8pt}
\caption{Quantitative comparison between using a globally consistent scale and the proposed SMEM. Values report the CD, with \textbf{bold} indicating the best results.} 
\centering
\begin{tabular}{c|cccc|cccc}
\toprule
                        & \multicolumn{4}{c|}{\textbf{Setting I, only regular views}}              & \multicolumn{4}{c}{\textbf{Setting II, regular + close-up views}}       \\
                       & Ganesha           & Lego            & Mic              & Ship             & Ganesha           & Lego            & Mic             & Ship             \\ \midrule
Consistent   scale     & 0.0653          & 0.1488          & 0.0845          & \textbf{0.0687} & 0.0541          & 0.1324          & 0.0909         & 0.0719          \\
SMEM                    & \textbf{0.0617} & \textbf{0.1474} & \textbf{0.0812} & 0.0704           & \textbf{0.0483} & \textbf{0.1131} & \textbf{0.0809} & \textbf{0.0684} \\ \bottomrule
\end{tabular}
\label{tab:sr}
\end{table*}

\para{Baselines.}
Based on high-fidelity 3D reconstruction task, we compare our method with the following categories:
\noindent (1) \textit{Multi-view normal integration:} \sn and \rnbneus are state-of-the-art (SOTA) approaches that take the same inputs and produce the same outputs as our method.
\noindent (2) \textit{Detailed surface reconstruction:} \petneus effectively captures high-frequency details from multi-view RGB images. For a fair comparison, we use both normal maps and RGB images for the training stage.
\noindent (3) \textit{Multi-scale representation:} Since the code for existing multi-scale representation–based 3D reconstruction methods~\cite{zhuang2023anti,ringnerf} is not publicly available, we choose \trimiprf which is designed for NVS and does not produce normal outputs. Hence, we train it using only RGB images. 

\para{Evaluation metrics.} 
We use Mean Angular Error (MAE) and L2 Chamfer Distance (CD) [mm] ~\cite{Kaya_Kumar_Oliveira_Ferrari_Gool, Knapitsch_Park_Zhou_Koltun_2017} to evaluate the accuracy of novel-view normal maps and 3D mesh, respectively. 


\begin{figure*}[t]
    \centering
    \includegraphics[width=0.9\linewidth]{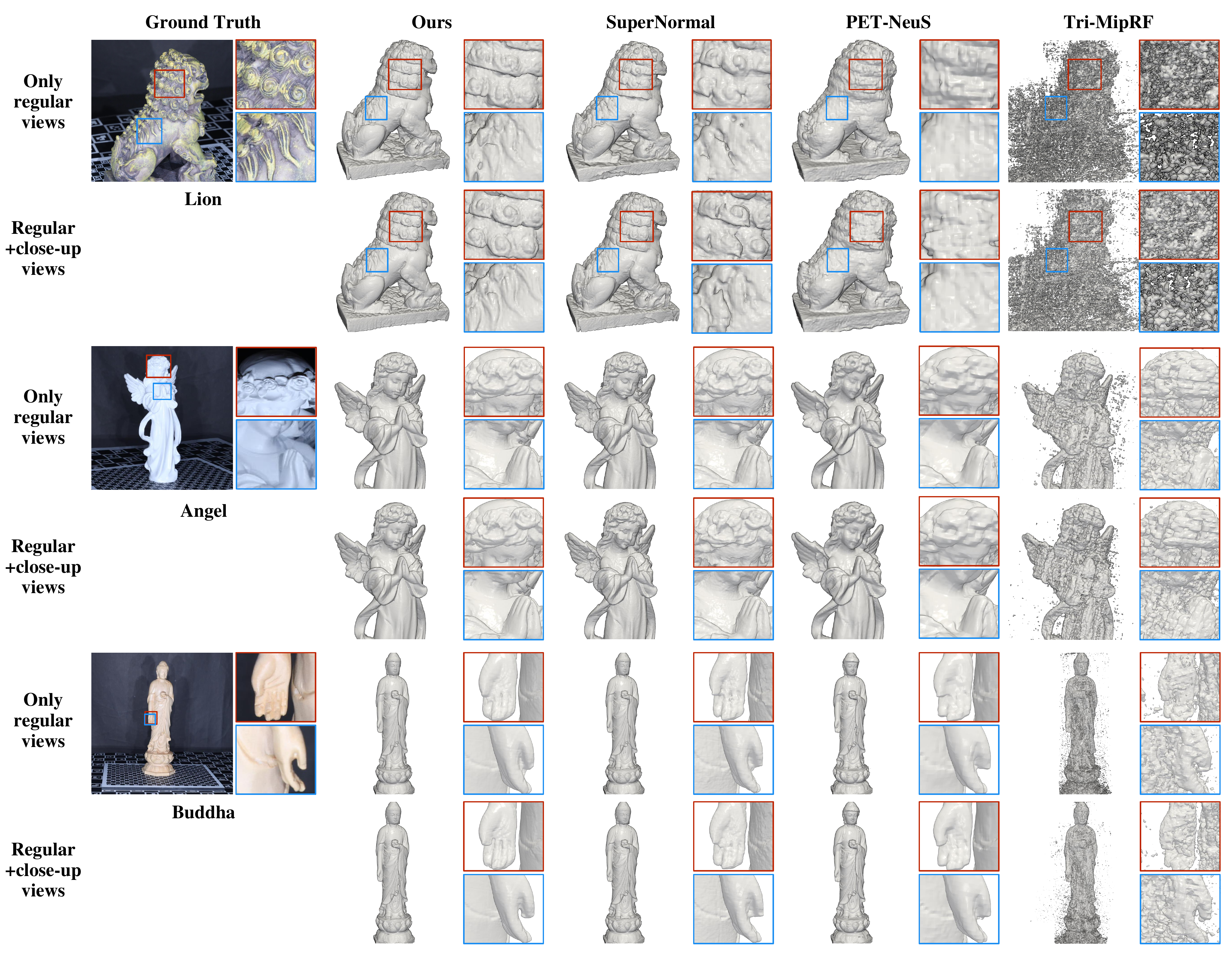}
    \caption{Comparison between our method and baselines on real-world data. (I) corresponds to Setting I, and (II) corresponds to Setting II. We zoom in on regions covered by close-up views and highly recommend that readers enlarge the figure to observe the differences clearly.}
    \label{fig:real}
\end{figure*}

\subsection{Comparison on synthetic data}
\label{subsec:exp1}
Table~\ref{tab:compare} shows MAE on six regular views and three close-up views, together with the CD for each object. Visualization results of \textit{Ganesha} and \textit{Lego} are shown in \fref{fig:compare}. Comparing the results of setting I and setting II, we observe that incorporating close-up views significantly improves the accuracy of 3D reconstruction, demonstrating the importance of such views for achieving high-quality geometry. However, introducing close-up views also increases scale variations among images and amplifies the influence of multi-view normal inconsistency. Our method explicitly accounts for the spatial coverage of each sampled point and takes the scale as one of the network inputs, leading to a multi-scale representation. This representation better preserves geometric details and mitigates multi-view normal inconsistency, producing renderings that are more consistent with ground truth. Consequently, our approach outperforms all baselines in both MAE and CD metrics.

In contrast, \rnbneus exhibits limited capability in capturing fine details, resulting in over-smoothed surfaces. Both \sn and \petneus adopt grid-based representations, which improve the modeling of high-frequency signals. However, their lack of multi-scale design makes them susceptible to noise in regions with large scale variations. \trimiprf projects the sampling scale onto multiple levels to estimate density. Although this constitutes a multi-scale representation, mesh extraction in \trimiprf suffers from two limitations. First, it can only extract meshes at a single scale, for which we use the finest level from the training stage. This is because \trimiprf lacks a mechanism to adaptively select the optimal scale for each spatial point, unlike our method which performs scale-adaptive mesh extraction.
Second, \trimiprf represents the scene using a density field, a soft representation that is inherently prone to noise.

For running time, we train each method on a single RTX A6000, with the training time displayed in the last column of Table~\ref{tab:compare}. \petneus takes the longest time due to its self-attention feature extractor, while ours runs slightly slower than \sn and \trimiprf because of additional scale encoding, yet achieves superior geometric accuracy.

\subsection{Comparison on real-world data}
\label{subsec:exp2}
For real-world experiments, we exclude \rnbneus, as it produces over-smoothed results on synthetic data and requires albedo inputs, which are difficult to obtain in real-world scenarios. As shown in \fref{fig:real}, our method reconstructs finer geometric details by effectively utilizing close-up views. By comparison, the meshes produced by other methods tend to degrade when close-up views are included. For instance, \sn exhibits lower accuracy on \textit{Angel}, whereas \petneus shows degraded performance on \textit{Lion}.  This degradation occurs because point-sampling-based methods exhibit increased multi-view normal inconsistency when close-up views are added. The noise of \trimiprf is exacerbated on the real-world dataset.

 \begin{figure}[t]
    \centering
    \includegraphics[width=\linewidth]{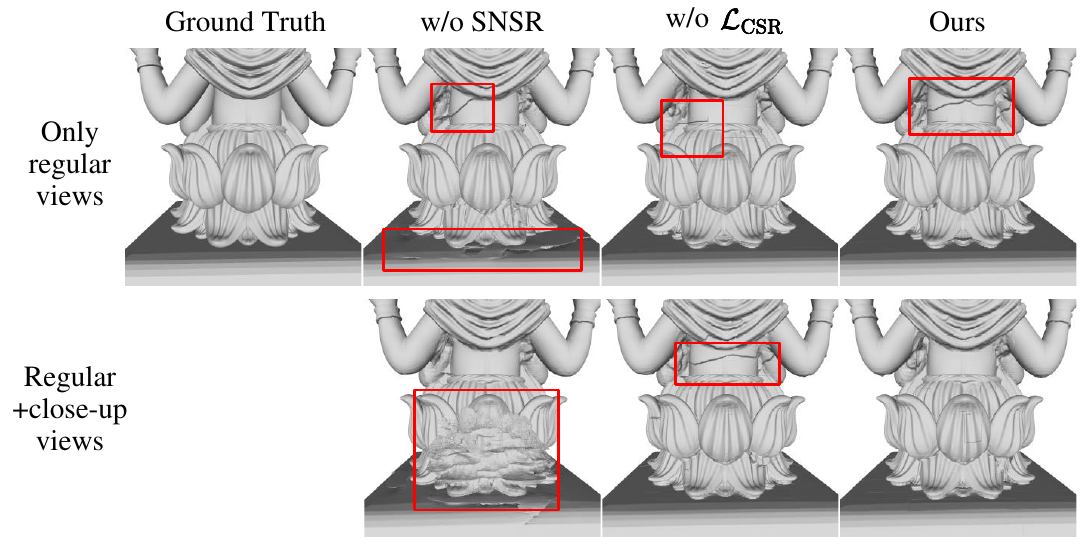}
    \caption{Qualitative comparisons on \textit{Ganesha} between our method and those without the proposed SNSR or $\mathcal{L}_\text{CSR}$.} 
    \label{fig:ablation1}
\end{figure}

\begin{figure}
    \centering
    \includegraphics[width=1\linewidth]{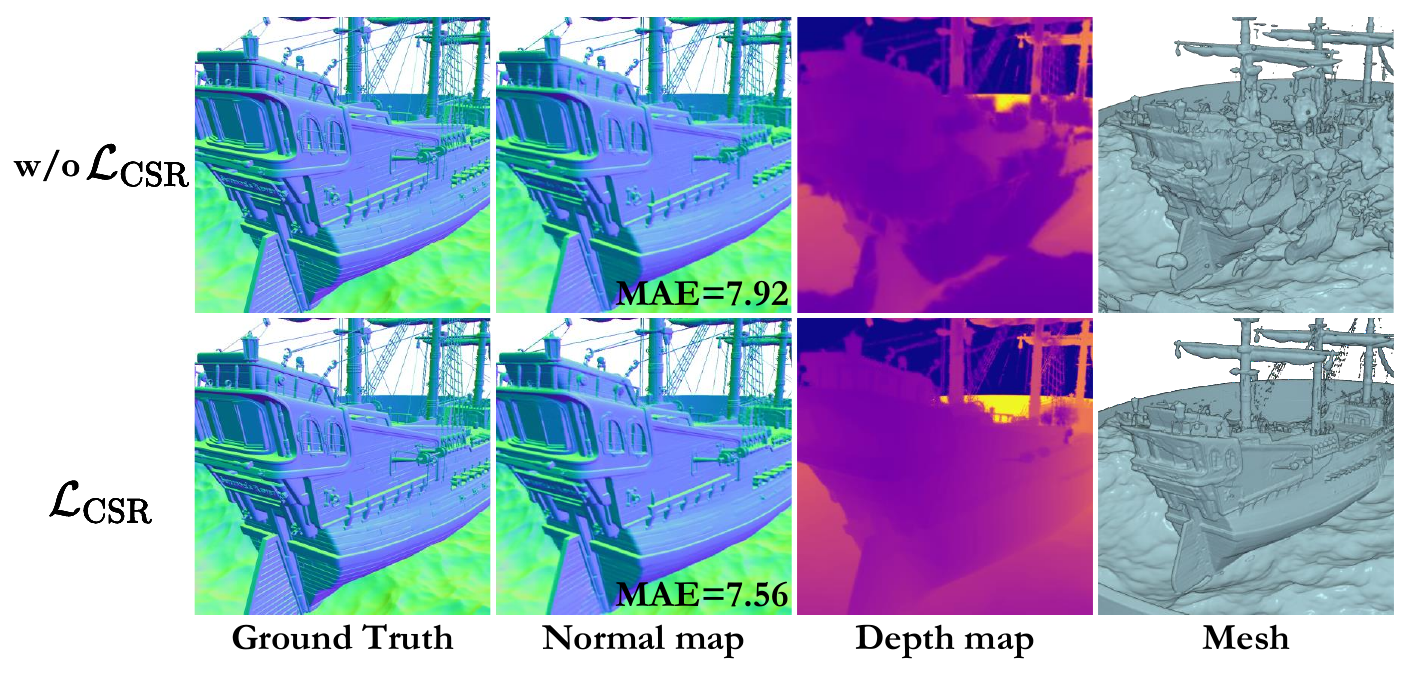}
    \caption{Qualitative comparison of the \textit{Ship} results with and without $\mathcal{L}_\text{CSR}$.}
    \label{fig:sdf}
\end{figure}

 \begin{figure}[t]
    \centering
    \includegraphics[width=0.9\linewidth]{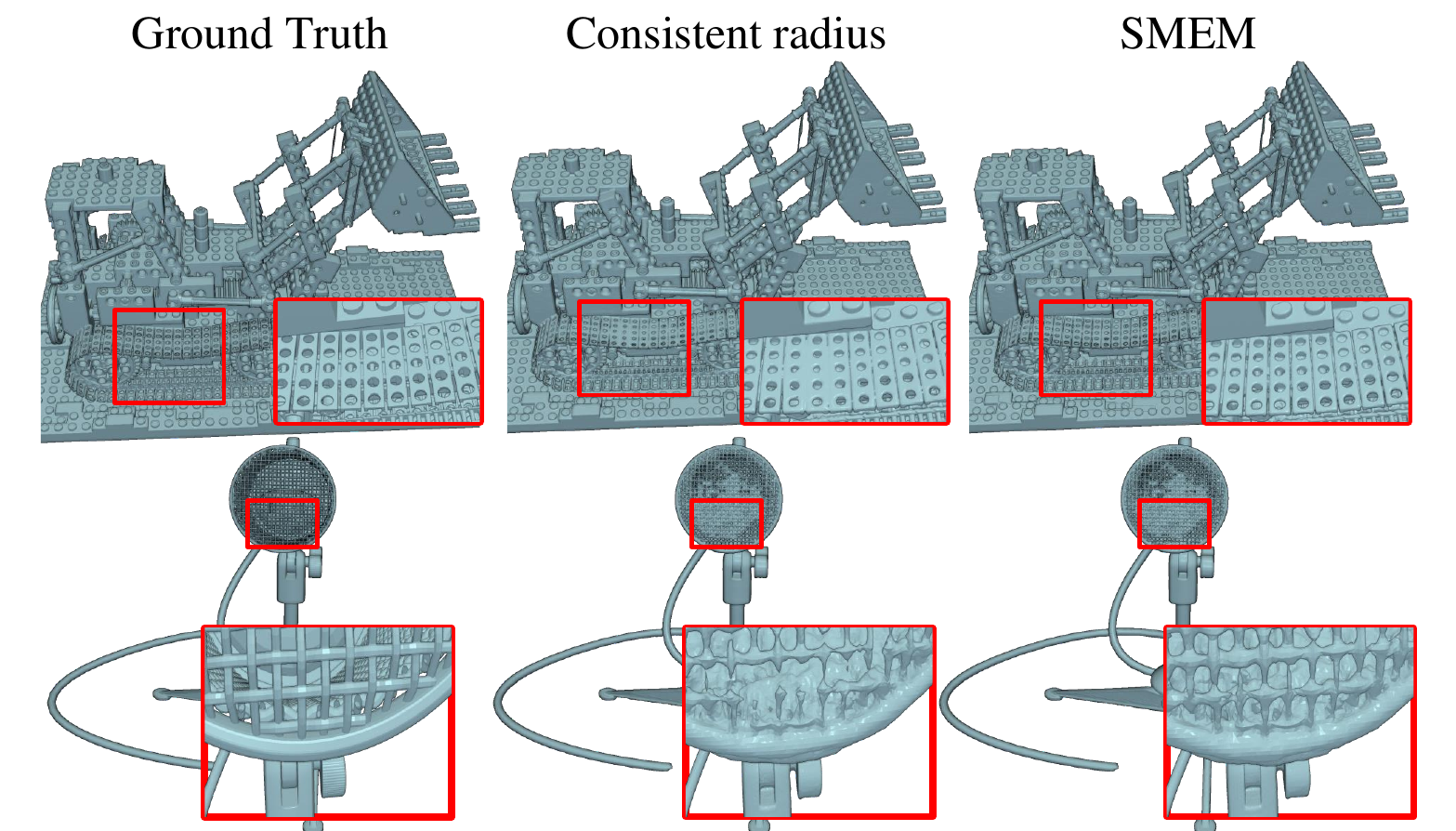}
    \caption{Qualitative comparisons of mesh extraction with and without SMEM in Setting II (regular + close-up views).} 
    \label{fig:ablation2}
\end{figure}

\subsection{Ablation study}
\label{subsec:ab}
Table~\ref{tab:ablation}, \fref{fig:ablation1} and \fref{fig:sdf}, demonstrate the effectiveness of both SNSR and $\mathcal{L}_{\text{CSR}}$. For SNSR, the improvement is more pronounced in Setting II (regular + close-up views). This is because Setting II involves various scales, and the presence of more scales tends to increase multi-view normal inconsistencies.
In contrast, in Setting I (only regular views), since the distance between the camera and the object remains constant, the sampling points share relatively uniform scales. As a result, although the effect is less noticeable, the performance still improves. The role of $\mathcal{L}_{\text{CSR}}$ is to enforce smoother meshes extracted from SNSR. \Fref{fig:ablation1} shows that without $\mathcal{L}_{\text{CSR}}$, the reconstructed meshes exhibit local non-smoothness. Additionally, \fref{fig:sdf} shows that the introduction of $\mathcal{L_{\text{CSR}}}$ helps prevent the model from overfitting.

Table~\ref{tab:sr} and \fref{fig:ablation2} illustrate the effectiveness of our proposed SMEM, which requires no additional computation. A straightforward alternative is to search for a globally consistent scale across all vertices over iterations, but this is highly time-consuming. For comparison, we manually set a suitable minimum and maximum scale, uniformly sampled $10$ scales within this range, and selected the best-performing mesh for evaluation. As evidenced by the lower CD in Table~\ref{tab:sr} and the finer details in \fref{fig:ablation2}, our method clearly outperforms this baseline in both efficiency and mesh quality.

\section{Conclusions}
In this paper, we propose SNSR, a novel multi-scale representation to address the multi-view normal inconsistency that arises from varying scales when normal maps are observed under different camera-to-object distances. Furthermore, we introduce SMEM, which adaptively generates a scale-dependent surface based on the observation. Experiments demonstrate that our method achieves high-fidelity surface reconstruction from normals under both regular and close-up settings. 

\subsection{Limitation}
Discontinuities appear on surfaces, such as the wings and face of \textit{Angel} in \fref{fig:real}. This issue may arise because our method relies on multi-resolution hash grids and analytical gradients. As a result, optimization updates propagate only within local hash grids, leading to a lack of non-local smoothness~\cite{li2023neuralangelo}. This problem could potentially be mitigated using the curvature loss proposed in~\cite{li2023neuralangelo}, but as it involves second-order derivatives, the computational cost increases significantly. We leave the exploration of this direction to future work.

\section*{Acknowledgments}
This work was supported the Beijing Key Laboratory of Multimodal Data Intelligent Perception and Governance, the Major Program of the National Natural Science Foundation of China (No. 62495085), the National Natural Science Foundation of China (No. 62442603, 62472044, U24B20155), the Key Research and Development Program of Liaoning Province (No. 2023JH26/10200014), the Beijing Major Science and Technology Project (No. Z251100007125021), the Hebei Natural Science Foundation Project (No. 242Q0101Z), the Beijing-Tianjin-Hebei Basic Research Funding Program (No. F2024502017), and the Program of China Scholarship Council (No. 202306060184).

\section*{Ethics Statement}
We did not conduct any experimental work with human participants or animals that would require an ethics approval or participant/patient consent.

\bibliographystyle{eg-alpha-doi} 
\bibliography{egbibsample}       



\end{document}